\documentclass[12pt,twoside]{article}

\usepackage{knvns98,amsthm,amsmath,ecltree,epic,eepic}

\allowdisplaybreaks

\newcommand{\Reals}{\mathrm{I} \! \mathrm{R}}
\newcommand{\Rn}{{\mathrm{I} \! \mathrm{R}}^n}
\newcommand{\Nats}{\mathrm{I} \! \mathrm{N}}

\newcommand{\Po}{\mbox{$\mathcal{P} \:$}}

\newcommand{\Rl}{\mbox{$\mathcal{R(L)} \:$}}
\newcommand{\La}{\mbox{$ \mathcal{L}\:$}}

\newtheorem{de}{Definition}

\title{Statistical Inference and Probabilistic Modeling for
Constraint-Based NLP}
 
\author{Stefan Riezler} 

\address{Seminar f\"ur Sprachwissenschaft, 
Universit\"at T\"ubingen \\
Wilhelmstra{\ss}e 113,
72074 T\"ubingen, Germany \\
email: riezler@sfs.nphil.uni-tuebingen.de}

\type{Lecture}

\contact{Stefan Riezler}

\begin{document}


\maketitle
 
\begin{abstract}
In this paper we present a probabilistic model for constraint-based
grammars and a method for estimating the
parameters of such models from incomplete, i.e., unparsed data. Whereas methods exist to estimate the parameters of probabilistic context-free
grammars from incomplete data (\cite{Baum:70}), so far for
probabilistic grammars involving context-dependencies only parameter
estimation techniques from complete, i.e., fully parsed data have been
presented (\cite{Abney:97}). However, complete-data estimation requires
labor-intensive, error-prone, and grammar-specific hand-annotating of
large language corpora. We present a log-linear probability model for
constraint logic programming, and a general algorithm to estimate the
parameters of such models from incomplete data by extending the
estimation algorithm of \cite{DDL:97} to incomplete
data settings.  
\end{abstract}
\begin{abstract}
Diese Arbeit pr\"asentiert ein probabilistisches Modell f\"ur
kontext-sensitive, constraint-basierte Grammatiken und erstmals eine
Methode, die Parameter solcher probabilistischer Modelle anhand
unvollst\"andiger Daten einzusch\"atzen. Probabilistische Grammatiken
werden hier in einem log-linearen Wahrscheinlichkeitsmodell f\"ur
Constraint Logik Programmierung formalisiert. Die pr\"asen-tierte
Parameter-Sch\"atzmethode ist eine Erweiterung des Improved Iterative
Scaling-Algorithmus von \cite{DDL:97} f\"ur Parametersch\"atzung
anhand unvollst\"andiger Daten. Die vorgestellten Methoden
erm\"oglichen die probabilistische Modellierung verschiedenster
constraint-basierter Grammatiken und ein automatisches Training solcher
Modelle anhand ungeparster Sprachdaten. 
\end{abstract}

\section{Introduction} 

Probabilistic grammars are of great interest for computational natural
language processing (NLP), e.g., because they allow the resolution of
structural ambiguities by a probabilistic ranking of competing
analyses. A prerequisite for such applications is parameter
estimation, i.e, a method to adapt the model parameters to best
account for a given language corpus. Clearly, an estimation technique
similar to the well-known maximization technique of \cite{Baum:70} for
context-free models would be desirable also for constraint-based
models. Baum's maximization technique permits model parameters to be
efficiently  estimated from incomplete, i.e., unparsed data rather than
from complete, i.e., fully parsed data.  
Recently, an attempt to apply this estimation technique to a
probabilistic version of the constraint logic programming (CLP) scheme of
\cite{HuS:88} has been presented by \cite{Eisele:94}. 
As recognized by Eisele, there is a
context-dependence problem associated with applying this technique to
constraint-based systems. That is, incompatible variable bindings 
can lead to failure derivations, which cause a loss of probability
mass in the estimated probability distribution over derivations. This
probability leakage prevents the estimation procedure from yielding
the desired maximum likelihood values in the general case. A similar
problem troubles every attempt to embed Baum's maximization technique
into an estimation procedure for probabilistic analogues of a
constraint-based processing systems (see, e.g., \cite{Briscoe:92},
\cite{BriscoeCarroll:93}, \cite{Brew:95}, \cite{Miyata:96}, or
\cite{Osborne:97})\footnote{Moreover, even approaches such as that of
\cite{Pereira:92}, where the derivation steps are in fact
context-free, must be characterized as constraint-based and
exhibit a similar problem because discarding derivations incompatible
with the bracketing of a training corpus from the estimation procedure
also induces a problem of loss of probability mass.}. From a mathematical
point of view, all such constraint-based approaches contradict the
inherent assumptions of Baum's maximization technique which require
that the derivation steps are mutually independent and that the set of
licensed derivations is unconstrained. Only recently, \cite{Abney:97} has shown how to overcome this problem by using the algorithm of
\cite{DDL:97} for estimation. This method, however, applies only to
complete data.

Unfortunately, the need to rely on large samples of complete
data is impractical. For parsing applications, complete data means several
person-years of hand-annotating large corpora with specialized
grammatical analyses. This task is always labor-intensive, error-prone,
and restricted to a specific grammar framework, a specific language,
and a specific language domain. Clearly, flexible techniques for
parameter estimation of probabilistic constraint-based 
grammars from \emph{incomplete data} are desirable.

The aim of this paper is to solve the problem of parameter estimation
from incomplete data for probabilistic constraint-based grammars.
For this aim, we present a log-linear probability model for CLP. CLP
is used here to provide an operational treatment of purely declarative
grammar frameworks such as PATR, LFG or HPSG\footnote{An example for
an embedding of feature-based constraint languages into the
CLP scheme of \cite{HuS:88} is the formalism CUF (\cite{Doerre:93}).}.
A probabilistic CLP scheme then yields a
formal basis for probabilistic versions of various constraint-based
grammar formalisms. The probabilistic model defines a probability
distribution over the proof trees of a constraint logic program on the
basis of weights assigned to arbitrary properties of the trees. In NLP
applications, such properties could be, e.g., simply context-free
rules or context-sensitive properties such as subtrees of proof trees
or non-local head-head relations. The algorithm we will present is an
extension of the estimation method for log-linear models of
\cite{DDL:97} to incomplete-data settings. Furthermore, we will
present a method for automatic property selection from incomplete data.

The rest of this paper is organized as follows. Section \ref{CLP}
introduces the basic formal concepts of CLP. Section
\ref{LogLinearModel} presents a log-linear model for probabilistic
CLP. Parameter estimation and property selection of log-linear models
from incomplete data is treated in
Sect. \ref{StatisticalInference}. Concluding remarks are made in 
Sect. \ref{Conclusion}.

\section{Constraint Logic Programming for NLP} \label{CLP}

In the following we will quickly report the basic concepts of the CLP
scheme of \cite{HuS:88}. A constraint-based grammar is encoded by a constraint
logic program \Po with constraints from a grammar constraint language
\La embedded into a relational programming constraint language \Rl. 

Let us consider a simple non-linguistic example. The program of
Fig. \ref{program} consists of five definite clauses with embedded 
\La-constraints from a language of hierarchical types. The ordering on the types is
defined by the operation of set inclusion on the denotations
($\cdot'$) of the types and $a' \subseteq c' \subseteq e'$, $b'
\subseteq d' \subseteq e'$, and $c' \cap d' = \emptyset$.

\begin{figure}[htbp]

{\footnotesize

\begin{center}
\begin{tabular}{l}
$s(Z) \leftarrow p(Z) \:\&\: q(Z).$ \\
$p(Z) \leftarrow Z=a.$ \\
$p(Z) \leftarrow Z=b.$ \\
$q(Z) \leftarrow Z=a.$ \\
$q(Z) \leftarrow Z=b.$ 
\end{tabular}
\end{center}

}

\caption{Simple constraint logic program}
\label{program}

\end{figure}

Seen from a parsing perspective, an input string corresponds to a
an initial goal or query $G$ which is a possibly empty conjunction of
\La-constraints and \Rl-atoms.  
Parses of a string (encoded by $G$) as produced by a grammar
(encoded by \Po) correspond to \Po-answers of $G$. A \Po-answer of a goal
$G$ is defined as a satisfiable \La-constraint
$ \phi$ s.t.\ the implication $\phi \rightarrow G$
is a logical consequence of \Po. 
The operational semantics of conventional logic programming,
SLD-resolution (\cite{Lloyd:87}), is generalized by performing goal
reduction only on the \Rl-atoms and solving conjunctions of collected
\La-constraints by a given \La-constraint solver. An example
for queries and proof trees for the program of Fig. \ref{program} is
given in Fig. \ref{prooftrees}.

In the following it will be convenient to view the
search space determined by this derivation procedure as a
search of a tree. Each derivation from a query $G$ and a program \Po
corresponds to a branch of a derivation tree, and each successful
derivation to a subtree of a derivation tree, called a proof tree,
with $G$ as root note and a \Po-answer as terminal node. 
We assume each parse of a sentence to be
associated with a single proof tree. In order to rank parses in terms of
their likelihood, we define a probability distribution over
proof trees. To this end we propose a log-linear model.

\section{A Log-Linear Probability Model for CLP} \label{LogLinearModel}

Log-linear models are powerful exponential probability distributions
which define the probability of an event as being proportional
to the product of weights assigned to selected properties of the
event\footnote{Log-linear models emerged in statistical physics as
  Gibbs- or Boltzmann-distribution and can be interpreted also as
  maximum entropy distributions \cite{Jaynes:57}.}.
For our application, the special instance of interest is a log-linear
distribution over the countably infinite set of proof trees for a set
of queries to a program. Log-linear distributions take the following form.

\begin{de} \label{LogLinearDistribution}
A log-linear probability distribution $p_\lambda$ on a set
  $\Omega$ is defined s.t.\ for all $\omega \in \Omega$:
\begin{center}
$p_\lambda(\omega) =
{Z_\lambda}^{-1}
e^{\lambda \cdot \nu(\omega)} p_0(\omega)$,
\end{center}

\begin{description}

\setlength{\parsep}{0ex}
\setlength{\itemsep}{0ex}

\item[]
$Z_\lambda = \sum_{\omega \in \Omega} 
e^{\lambda \cdot \nu(\omega)} p_0(\omega)$
is a normalizing constant,
\item[] $\lambda = (\lambda_1, \ldots, \lambda_n) \in \Rn$
is a vector of log-parameters,
\item[] $\mathbf{\nu} = (\nu_1, \ldots, \nu_n)$
is a vector of property-functions s.t. for each 
$\nu_i:\Omega
\rightarrow \Nats $, 
$\nu_i(\omega)$
is the number of occurences of the $i$-th property in $\omega$, 
\item[]$\lambda \cdot \nu (\omega)$
is a weighted property-function s.t. 
$\lambda \cdot \nu (\omega) =
\sum^n_{i=1}\lambda_i \nu_i(\omega)$,
\item[]$p_0$ is a fixed initial distribution.
\end{description}
\end{de}

When we search for a proper probability distribution over given
training data in a maximum likelihood estimation framework, we want to
find a distribution reflecting the statistics of the
training corpus. This means, we have to choose useful properties
(property selection) and appropriate corresponding log-parameters
(parameter estimation). A definition of properties convenient for our
application is as subtrees of proof trees. 

\begin{figure}[htbp]
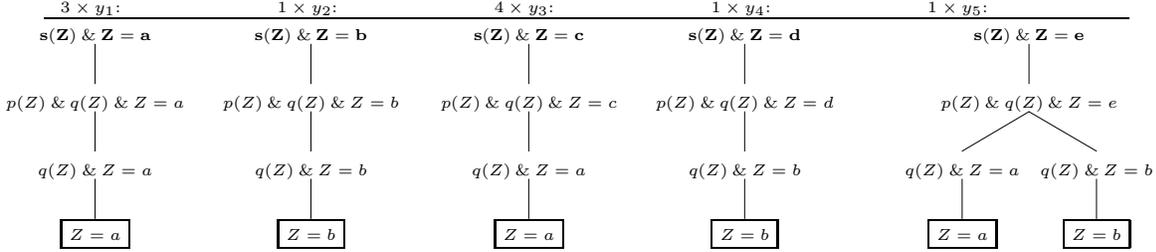

{\tiny

\begin{center} 
\begin{tabular}{p{70pt}p{70pt}p{70pt}p{70pt}p{70pt}}

$3 \times y_1$: & $1 \times y_2$: & $4 \times y_3$:
& $1 \times y_4$: & $1 \times y_5$: \\ \hline

\begin{bundle}{$\mathbf{s(Z) \:\&\: Z=a}$}
\chunk
{
\begin{bundle}{$p(Z) \:\&\: q(Z) \:\&\: Z=a$}
\chunk
{
\begin{bundle}{$q(Z) \:\&\: Z=a$}
\chunk
{\fbox{$Z=a$}}
\end{bundle}
}
\end{bundle}
}
\end{bundle}

& 

\begin{bundle}{$\mathbf{s(Z) \:\&\: Z=b}$}
\chunk
{
\begin{bundle}{$p(Z) \:\&\: q(Z) \:\&\: Z=b$}
\chunk
{
\begin{bundle}{$q(Z) \:\&\: Z=b$}
\chunk
{\fbox{$Z=b$}}
\end{bundle}
}
\end{bundle}
}
\end{bundle}
 
&

\begin{bundle}{$\mathbf{s(Z) \:\&\: Z=c}$}
\chunk
{
\begin{bundle}{$p(Z) \:\&\: q(Z) \:\&\: Z=c$}
\chunk
{
\begin{bundle}{$q(Z) \:\&\: Z=a$}
\chunk
{\fbox{$Z=a$}}
\end{bundle}
}
\end{bundle}
}
\end{bundle}

&

\begin{bundle}{$\mathbf{s(Z) \:\&\: Z=d}$}
\chunk
{
\begin{bundle}{$p(Z) \:\&\: q(Z) \:\&\: Z=d$}
\chunk
{
\begin{bundle}{$q(Z) \:\&\: Z=b$}
\chunk
{\fbox{$Z=b$}}
\end{bundle}
}
\end{bundle}
}
\end{bundle}

&

\setlength{\GapWidth}{25pt}
\begin{bundle}{$\mathbf{s(Z) \:\&\: Z=e}$}
\chunk
{
\begin{bundle}{$p(Z) \:\&\: q(Z) \:\&\: Z=e$}

\chunk
{
\begin{bundle}{$q(Z) \:\&\: Z=a$}
\chunk
{\fbox{$Z=a$}}
\end{bundle}
}

\chunk
{
\begin{bundle}{$q(Z) \:\&\: Z=b$}
\chunk
{\fbox{$Z=b$}}
\end{bundle}
}

\end{bundle}
}
\end{bundle}

\end{tabular}

\end{center}

\caption{Queries and proof trees for constraint logic program}
\label{prooftrees}
}
\end{figure}

Suppose we have a training corpus of ten queries, consisting of three
tokens of query $y_1: \: \mathbf{s(Z)\:\&\: Z=a}$, four tokens of
$y_3: \: \mathbf{s(Z)\:\&\: Z=c}$, and one token each of query
$y_2:\: \mathbf{s(Z)\:\&\: Z=b},
\; y_4: \: \mathbf{s(Z)\:\&\: Z=d}$, and
$y_5: \: \mathbf{s(Z)\:\&\: Z=e}$.
The corresponding proof trees generated by the program in
Fig. \ref{program} are given in Fig. \ref{prooftrees}. Note that
queries $y_1$, $y_2$, $y_3$ and $y_4$ are unambiguous, being assigned
a single proof tree, while $y_5$ is ambiguous.

A first useful distinction between the proof trees of
Fig. \ref{prooftrees} can be obtained by
selecting the two subtrees $t_1: \fbox{$Z=a$}$ and $t_2:
\fbox{$Z=b$}$ as properties. This allows us to cluster the
proof trees into two disjoint sets on the basis of their similar statistical
qualities.
Since in our training corpus seven out of ten queries
come unambiguously with a proof tree including property $t_1$, we
would expect the maximum likelihood parameter value corresponding to
property $t_1$ to be higher than the parameter value of property
$t_2$. However, we cannot simply recreate the proportions of the
training data from the corresponding proof trees because we do not
know the frequency of the possible proof trees of query $y_5$. A
solution to this incomplete-data problem is presented in the next section.

\section{Inducing Log-Linear Models from Incomplete Data}
\label{StatisticalInference}

As shown in the foregoing example, statistical inference for log-linear
models involves two problems: parameter estimation and property
selection. In the following, we will present the details of an
algorithm to solve to these problems in the presence of incomplete data.

\subsection{Parameter Estimation}

The ``improved iterative scaling'' algorithm presented by \cite{DDL:97} solves a
maximum likelihood estimation problem for log-linear models with respect to complete
data\footnote{For complete data, this is equivalent to solving a
  constrained maximum entropy problem.}. This algorithm itself is an
extension of the ``generalized iterative scaling'' algorithm of
\cite{Darroch:72} especially tailored to estimating models with large
parameter spaces. We present a version of the first algorithm
specifically designed for incomplete data problems. A proof of
monotonicity and convergence of the algorithm is given in the
appendix, i.e, we show that succesive steps of the algorithm increase
the incomplete-data log-likelihood and eventually lead to convergence
to a (local) maximum.

Applying an incomplete-data
framework to a log-linear probability model for CLP, we can assume the
following to be given:

\begin{itemize}
\item observed, incomplete data $y \in \mathcal{Y}$,
  corresponding to a given, finite sample of queries for a
  constraint logic program \Po, 
\item unobserved, complete data $x \in \mathcal{X}$, corresponding to
  the countably infinite sample of proof trees for queries $\mathcal{Y}$
  from \Po,
\item a many-to-one function
$Y:\mathcal{X} \rightarrow \mathcal{Y}$ 
  s.t. $Y(x) = y$ corresponds to the unique query labeling proof tree
  $x$, and its inverse
  $X:\mathcal{Y} \rightarrow \mathcal{X}$
  s.t. $X(y) = \{x |\; Y(x) = y \}$
  is the countably infinite set of proof trees for query $y$ from \Po,
\item a complete-data specification $p_\lambda (x)$, which is a
  log-linear distribution on $\mathcal{X}$ 
  with given initial distribution $p_0$,
  fixed property-functions vector $\nu$,
  and depending on parameter vector $\lambda$, 
\item an incomplete-data specification $p_\lambda (y)$, which is
  related to the complete-data specification by
  $p_\lambda(y) = \sum_{x \in X(y)} p_\lambda(x).$
\end{itemize}
For the rest of this section we will refer to a given vector $\nu$ of
property functions, which is assumed to result from the property
selection procedure presented below.
If the incomplete-data log-likelihood function $L$ is defined over a 
sample $\mathcal{Y}$ of tokens of queries $y$ s.t.
$L(\lambda)=\ln \prod_{y \in \mathcal{Y}} p_\lambda (y)$,
then the problem of maximum-likelihood estimation for log-linear
models from incomplete data can be stated as follows.
  Given a fixed $\mathcal{Y}$-sample and a set $\Lambda =
  \{ \lambda |\; p_{\lambda}(x)$ is a log-linear distribution on
  $\mathcal{X}$ with fixed $p_0$, fixed $\nu$ and
  $\lambda \in \Rn \}$,
  we want to find a maximum likelihood estimate
  $\lambda^\ast$ of $\lambda$ s.t.
  $\lambda^\ast = {\arg\max}_{\lambda \in \Lambda} \;
  L(\lambda)$.

The key idea of the proposed method is to iteratively maximize a
strictly concave auxiliary function when the non-concave
log-likelihood function cannot be maximized analytically.
Following \cite{DDL:97},
we can define an auxiliary function $A$ directly as a lower bound
on $L(\gamma +\lambda) - L(\lambda)$,
i.e., as a conservative estimate of the difference in log-likelihood
when going from a basic model $p_\lambda$ to an extended model
$p_{\gamma + \lambda}$\footnote{Another possibility to arrive
  at the same auxiliary function is to use the complete-data auxiliary
  function of \cite{DDL:97} in the M-step of a generalized EM
  algorithm \cite{Dempster:77}. This guarantees
  monotonicity of the resulting algorithm, but convergence yet has to
  be proved. Our approach views the incomplete-data auxiliary function
  directly as a lower bound on the improvement in incomplete-data
  log-likelihood, which enables an intuitive and elegant proof of convergence.}.
The specific design of $A$ for incomplete data can be derived from the
complete data case, in essence, by replacing an expectation of
complete, but  unobserved, data by a conditional expectation given the
observed data and the current parameter
values\footnote{$k_{\lambda}(x|y) = p_{\lambda}(x) / \sum_{x \in X(y)}
  p_{\lambda}(x) $  is the conditional probability of the complete data
  $x$ given the observed data $y$ and the current fit of the parameter
  $\lambda$.
  Furthermore, $p[f] = \sum_{\omega \in \Omega} p(\omega)
  f(\omega)$ is the expectation of a function
  $f:\Omega \rightarrow \Reals$ with respect to a probability distribution
  $p$ on a set $\Omega$, 
  $\nu_{\#}(x) = \sum^n_{i=1} \nu_i(x)$,
  ${\bar{\nu}}_i(x) = \nu_i(x) / \nu_{\#}(x)$.}.
Let  $\lambda \in \Lambda$, $\gamma \in \Rn$. Then
\begin{center}
$A(\gamma , \lambda) = \sum_{y \in \mathcal{Y}} 
( 1 + k_{\lambda} \left[ \gamma \cdot \nu \right] -
p_{\lambda} \left[ \sum^n_{i=1} {\bar{\nu}}_i e^{\gamma_i \nu_{\#} }\right] 
)$.
\end{center}

$A$ is maximized in $\gamma$ at the unique point $\hat\gamma$ satisfying for each $\hat\gamma_i$:
\begin{center}
$\sum_{y \in \mathcal{Y}}  k_{\lambda} \left[ \nu_i \right] 
= \sum_{y \in \mathcal{Y}} p_{\lambda} \left[ \nu_i
    e^{\hat\gamma_i \nu_{\#} }\right] .$
\end{center}

An iterative algorithm for maximizing $L$ is constructed from $A$ as
follows. For the want of a name, we will call this the ``Iterative
Maximization (IM)'' algorithm.

\begin{de}[Iterative maximization] \label{IterativeMaximization}
Let $\mathcal{M}:\Lambda
  \rightarrow \Lambda$ be a mapping defined by
$
\mathcal{M}(\lambda) = \hat \gamma + \lambda
\textrm{ with }
\hat \gamma = {\arg\max}_{\gamma \in \Rn} \;
A(\gamma , \lambda).
$
\textrm{ Then each step of the IM algorithm is
  defined by }
$
\lambda^{(k+1)} = \mathcal{M}(\lambda^{(k)}).
$
\end{de}

As shown in the appendix, this procedure stepwise increases the
log-likelihood function $L$ and eventually converges to a (local)
maximum of $L$. For large configuration spaces $\mathcal{X}$ the
expectations to be calculated can get intractable. Here approximations
by conditional models or Monte Carlo methods have to be used.

\subsection{Property Selection}

A further problem is that exhaustive sets of properties can get
unmanageably large. Let properties of proof trees be defined as
connected subgraphs of proof trees, and suppose that properties can
incrementally be constructed by selecting from an initial set of goals
and from subtrees built by performing a resolution step at a terminal node of
a subtree already in the model.
Clearly, the exponentially growing set of possible properties must be
pruned by some quality measure. An appropriate measure can then be
used to define an algorithm for automatic property selection.

A straightforward  measure to take would be the improvement in
log-likelihood when extending a model by a single candidate property
$c$ with corresponding parameter $\alpha$. This would require iterative
maximization for each candidate property and is thus infeasible.
Following \cite{DDL:97}, we could  instead approximate the
improvement due to adding a single property by adjusting only the parameter of
this candidate and holding all other parameters of the model
fixed. Unfortunately, the incomplete-data log-likelihood $L$ is not
concave in the parameters and thus cannot be maximized directly.
However, we can instantiate the auxiliary function $A$ used in
parameter estimation to the extension of a
model $p_\lambda$ by a single property $c$ with log-parameter
$\alpha$,
i.e., we can express an approximate gain
$G_c(\alpha , \lambda)$ 
of adding a candidate property $c$ with log-parameter value $\alpha$
to a log-linear model $p_\lambda$ as a conservative estimate of the
true gain in log-likelihood as follows.

\begin{center}
$ G_c(\alpha , \lambda) 
= \sum_{y \in \mathcal{Y}} (
1 + 
k_\lambda [\alpha c] - 
p_\lambda[e^{\alpha c}] )$.
\end{center}

$ G_c(\alpha , \lambda) $ is maximized in $\alpha$
at the unique point $\hat\alpha$ satisfying
\begin{center}
$\sum_{y \in \mathcal{Y}}
k_\lambda [c] 
= 
\sum_{y \in \mathcal{Y}}
p_\lambda [ c \: e^{\hat\alpha c} ]$.
\end{center}

Property selection will incorporate that property out of the set
of candidates that gives the greatest improvement to the model at the property's
best adjusted parameter value. Since we are interested only in
relative, not absolute gains, a single, non-iterative maximization of
the approximate gain will be sufficient to choose from the candidates.

\subsection{Combined Statistical Inference}

A combined algorithm for statistical inference for log-linear models
from incomplete data is as follows.

\begin{description}
\item[{Input}] Initial model $p_0$, multiple incomplete-data sample
  $\mathcal{Y}$.

\item[{Output}] Log-linear model $p^\ast$ on complete-data sample
$\mathcal{X}$
with selected property function vector $\nu^\ast$ and 
estimated log-parameter vector $\lambda^\ast = {\arg\max\;}_{\lambda
\in \Lambda} \;  L(\lambda) $  
where $\Lambda = \{ \lambda | \; p_\lambda$ 
is a log-linear model on $\mathcal{X}$ 
based on $p_0$, $\nu^\ast$ and $\lambda \in \Rn \} $.

\item[Procedure] \mbox{}

\begin{enumerate}
\item $p^{(0)} := p_0$ with $C^{(0)} := \emptyset$,

\item Property selection: For each candidate property $c \in C^{(t)}$,
compute the gain 
$G_c(\lambda^{(t)}) := {\max}_{\alpha \in \Reals} \;
G_c(\alpha , \lambda^{(t)})$, 
and select the property $\hat c := {\arg\max}_{c \in
C^{(t)}} \;
G_c(\lambda^{(t)})$.

\item Parameter estimation: Compute a maximum likelihood parameter
  value 
$\hat \lambda := {\arg\max}_{\lambda\in\Lambda} \; L(\lambda)$ 
where $\Lambda = \{ \lambda | \; p_\lambda (x) $
is a log-linear distribution on $\mathcal{X}$
with initial model $p_0$,
property function vector $\hat \nu := \nu^{(t)} + \hat c$, 
and $\lambda \in \Rn \} $.

\item Until the model converges, set \\
$p^{(t+1)} := p_{\hat\lambda \cdot \hat\nu}$, 
$t := t+1$, 
go to $2$.

\end{enumerate}
\end{description}

Let us return to the example of Sect. \ref{LogLinearModel} and apply
the IM algorithm to the incomplete-data problem stated there.
For the selected properties $t_1$ and $t_2$, we have
$\nu_\#(x) = \nu_1(x) + \nu_2(x) = 1$
for all possible proof trees $x$ 
for the sample of Fig. \ref{prooftrees}. Thus the parameter updates
$\hat\gamma_i$ can be calculated from a particularly simple closed
form as follows.
\[
\hat\gamma_i = \ln \frac{\sum_y k_\lambda [\nu_i]}
{\sum_y p_\lambda [\nu_i]}.
\]
A sequence of IM iterates up to stability in the third
place after the decimal point of the incomplete-data log-likelihood is
given in Table \ref{IMestimation}. 
Probabilities of proof trees involving property $t_i$ are
denoted by $p_i$. Starting from an initial uniform probability of
$1/6$ for each proof tree, this estimation sequence converges to the
desrided accuracy after three iterations and yield probabilities $p_1
\approx .259$ and $p_2 \approx .074$ for the respective proof trees.

\begin{table}[htbp]
{\footnotesize
\begin{center}
\begin{tabular}{c|l|l|l|l|c}
Iteration $t$& $\lambda_1^{(t)}$ & $\lambda_2^{(t)}$ & $p_1^{(t)}$ & $p_2^{(t)}$ & $L(\lambda^{(t)})$ \\
\hline

0 & 0 & 0 & $1/6$ & $1/6$ & $-17.224448$ \\ \hline

1 & $\ln 1.5$ & $\ln .5$ & $.25$ & $.08\dot{3}$ & $-15.772486$ \\
\hline

2 & $\ln 1.55$ & $\ln .45$ & $.258\dot{3}$ & $.075$ & $-15.753678$ \\
\hline

3 & $\ln 1.555$ & $\ln .445$ & $.2591\dot{6}$ & $.0741\dot{6}$
 &
$-15.753481$ 
\end{tabular}
\end{center}
}
\caption{Estimation using the IM algorithm}
\label{IMestimation}
\end{table}

\section{Conclusion}
\label{Conclusion}

We have presented a probabilistic model for CLP, coupled with an algorithm
to induce the parameters and properties of log-linear models
from incomplete data. This algorithm is applicable to log-linear
probability distributions in general, and has been shown here to be
useful to estimate the parameters of probabilistic context-sensitive
NLP models. In contrast to related
approaches such as that of \cite{Abney:97} or
\cite{Magerman:94}, our statistical inference
algorithm provides the means for automatic and reusable training of
probabilistic constraint-based grammars from unparsed
corpora.

Furthermore, heuristic search algorithms for finding the
most probable analysis in the CLP model can be based upon this
probability model. For example, a combination of the
dynamic-programming techniques of Earley deduction \cite{Pereira:83} and
Viterbi-searching \cite{Viterbi:67} could be employed. Depending on
the class of constraint-based grammars under consideration, a
considerable gain in search efficiency can be
obtained\footnote{However, note that the choice of a particular class of constraint-based grammars also
  influences the behaviour of the algorithm in finding the optimal
  analysis. For example, in grammars where variable-bindings are
  ignored in Viterbi-searching in order to avoid the overhead of storing
  each variable binding separately the problem of pursuing a
  non-optimal path arises. In such cases only approximate heuristic
  searching can be done (see \cite{CarrollBriscoe:92} for a similar
  approach to a Viterbi-like heuristic search procedure for
  unification-based grammars).}.

The statistical inference algorithm presented is fully implemented and
has already been tested empirically with simple examples. Clearly, the
performance of the presented techniques in real-world NLP problems has
to be thoroughly investigated. Unfortunately, the current availability
of broad-coverage constraint-based grammars limited so far the
empirical evaluation of the presented techniques for the area of
constraint-based parsing. In future work we also will investigate the
applicability of the statistical methods here described to NLP problems other than constraint-based parsing.

{\footnotesize

\begin{appendix}

\section{Iterative Maximization: Propositions and Proofs}

In the following, we assume that for each property function $\nu_i$
some proof tree $x \in \mathcal{X}$ with $\nu_i(x) > 0$ exists,
and require $p_\lambda$ to be strictly positive on $\mathcal{X}$,
i.e., $p_\lambda(x) > 0$ for all $x \in \mathcal{X}$.
Furthermore, $p_{\gamma + \lambda}(x) = 
{Z_{\gamma \circ \lambda}}^{-1}
e^{\gamma \cdot \nu(x)} 
p_\lambda (x)$   
denotes an extended log-linear model with
$Z_{\gamma \circ \lambda} = p_\lambda[e^{\gamma \cdot
  \nu}]$. 

Lemma \ref{A<L-L} shows that the auxiliary function
$A(\gamma , \lambda)$
is a lower bound on the incomplete-data log-likelihood difference
$L(\gamma + \lambda) - L(\lambda)$.

\newtheorem{lemma}[po]{Lemma}
\begin{lemma} \label{A<L-L}
$A(\gamma , \lambda) \leq L(\gamma + \lambda) - L(\lambda)$.
\end{lemma}

{\footnotesize
\begin{proof}
\begin{eqnarray*}
L(\gamma + \lambda) - L(\lambda)
& = &
\sum_{y \in \mathcal{Y}} (
\ln \frac{p_{\gamma +\lambda}(y)}{p_{\lambda}(y)}
) \\
& = & 
\sum_{y \in \mathcal{Y}} (
\ln \frac{1}{p_{\lambda}(y)} 
\sum_{x \in X(y)} (
p_{\gamma + \lambda}(x) 
\frac{p_{\lambda}(x)}{p_{\lambda}(x)}
)
) \\
& = & 
\sum_{y \in \mathcal{Y}} (
\ln \sum_{x \in X(y)} (
\frac{p_{\lambda}(x)}{p_{\lambda}(y)}
\frac{p_{\gamma + \lambda}(x)}{p_{\lambda}(x)}
)
) \\
& \geq & 
\sum_{y \in \mathcal{Y}} (
\sum_{x \in X(y)} (
\frac{p_{\lambda}(x)}{p_{\lambda}(y)}
\ln \frac{p_{\gamma + \lambda}(x)}{p_{\lambda}(x)}
)
) 
\textrm{ by Jensen's inequality} \\
& = & 
\sum_{y \in \mathcal{Y}} (
\sum_{x \in X(y)} (
\frac{p_{\lambda}(x)}{p_{\lambda}(y)} (
\ln p_{\gamma + \lambda}(x) 
- \ln p_{\lambda}(x)
)
)
)\\
& = & 
\sum_{y \in \mathcal{Y}} (
\sum_{x \in X(y)} (
\frac{p_{\lambda}(x)}{p_{\lambda}(y)} (
\ln Z_{\gamma \circ \lambda}^{-1}
+ \ln e^{\gamma \cdot \nu(x)} 
+ \ln p_{\lambda}(x)
- \ln p_{\lambda}(x)
))) \\
& = & 
\sum_{y \in \mathcal{Y}} (
k_{\lambda} [ \gamma \cdot \nu ]
- \ln p_{\lambda} [ e^{\gamma \cdot \nu} ]
)\\
& \geq & 
\sum_{y \in \mathcal{Y}} (
k_{\lambda} [ \gamma \cdot \nu ]
+1 
-  p_{\lambda} [ e^{\gamma \cdot \nu} ]
) \quad \textrm{since } \ln x \leq x -1 \\
& = & 
\sum_{y \in \mathcal{Y}} (
k_{\lambda} [ \gamma \cdot \nu ]
+1 
- \sum_{x \in \mathcal{X}} (
p_{\lambda} (x) e^{ \sum^n_{i=1} \gamma_i \nu_i(x)
  \frac{\nu_\#(x)}{\nu_\#(x)}}
)
)\\
& = & 
\sum_{y \in \mathcal{Y}} (
k_{\lambda} [ \gamma \cdot \nu ]
+1 
- \sum_{x \in \mathcal{X}} (
p_{\lambda} (x) e^{ \sum^n_{i=1} \gamma_i \bar\nu_i(x) \nu_\#(x)}
)
)\\
& \geq &
\sum_{y \in \mathcal{Y}} (
k_{\lambda} [ \gamma \cdot \nu ]
+1 
- \sum_{x \in \mathcal{X}} (
p_{\lambda} (x) \sum^n_{i=1} \bar\nu_i(x) e^{  \gamma_i \nu_\#(x)}
)
) 
\textrm{ by Jensen's inequality}
\\
& = & 
\sum_{y \in \mathcal{Y}} (
k_{\lambda} [ \gamma \cdot \nu ]
+1 
- p_{\lambda} [ \sum^n_{i=1} \bar\nu_ie^{  \gamma_i \nu_\#}
]
) \\
& = & 
A(\gamma , \lambda).
\qed
\end{eqnarray*}
\renewcommand{\qed}{}
\end{proof}
}
Lemma \ref{A0=0} shows that there is no estimated improvement in
log-likelihood at the origin, and Lemma \ref{dA=dL} shows that the
critical points of interest are the same for $A$ and $L$.

\begin{lemma} \label{A0=0}
$A(0,\lambda) = 0$. 
\end{lemma}

\begin{lemma} \label{dA=dL}
$ \left. \frac{d}{dt} \right|_{t=0} A(t\gamma , \lambda)
= \left. \frac{d}{dt} \right|_{t=0} L(t\gamma + \lambda)$.
\end{lemma}
Theorem \ref{IncreasingIMLikelihood} shows the monotonicity of the IM
algorithm.

\newtheorem{theorem}[po]{Theorem}
\begin{theorem} \label{IncreasingIMLikelihood}
For all $\lambda \in \Lambda$:
$L(\mathcal{M}(\lambda) )\geq L(\lambda)$ with equality iff $\lambda$
is a fixed point of $\mathcal{M}$ or equivalently is a critical point of
$L$.
\end{theorem}

{\footnotesize
\begin{proof}
\begin{eqnarray*}
L(\mathcal{M}(\lambda)) - L(\lambda) 
& \geq & 
A(\mathcal{M}(\lambda)) \quad \textrm{by Lemma \ref{A<L-L}} \\
& \geq &
0 \quad \textrm{by Lemma \ref{A0=0} and definition of $\mathcal{M}$}.
\end{eqnarray*}
The equality $L(\mathcal{M}(\lambda)) = L(\lambda)$  holds iff
$\lambda$ is a fixed point of $\mathcal{M}$,
i.e., $\mathcal{M}(\lambda) = \hat\gamma + \lambda$ with
$\hat\gamma = 0$.
Furthermore, $\lambda$ is a fixed point of $\mathcal{M}$ iff 
$\hat\gamma = \underset{\gamma \in \Rn}{\arg\max}\; A(\gamma ,
\lambda) = 0$, \\
$\iff \textrm{for all } \gamma \in \Rn : \hat t = \underset{t \in
  \Reals }{\arg\max} \; A(t\gamma ,
\lambda) = 0$, \\
$\iff \textrm{for all } \gamma \in \Rn: \left. \frac{d}{dt} \right|_{t=0} A(t
\gamma , \lambda) = 0$, \\
$\iff \textrm{for all }\gamma \in \Rn: \left. \frac{d}{dt} \right|_{t=0}
L(t\gamma + \lambda) = 0 $, by Lemma \ref{dA=dL} \\
$\iff  \lambda \textrm{ is a critical point of $L$}.$
\end{proof}
}
Corollary \ref{MaximumLikelihoodEstimates} implies that a maximum
likelihood estimate is a fixed point of the mapping $\mathcal{M}$.

\newtheorem{corollary}[po]{Corollary}
\begin{corollary} \label{MaximumLikelihoodEstimates}
Let $\lambda^\ast = \underset{\lambda \in \Lambda}{\arg\max}\; L(\lambda)$.
Then $\lambda^\ast$ is a fixed point of $\mathcal{M}$.
\end{corollary}
Theorem \ref{convergence} discusses the convergence properties of the
IM algorithm. In contrast to the improved iterative scaling algorithm,
we cannot show convergence to a global maximum of a strictly concave
objective function. Rather, similar to the EM algorithm, we can show
convergence of a sequence of IM iterates to a critical point of the non-concave
incomplete-data log-likelihood function $L$.

\begin{theorem}[Convergence]\label{convergence}
Let $\{ \lambda^{(k)} \}$ be a sequence in $\Lambda$
determined by the IM Algorithm.
Then all limit points of $\{ \lambda^{(k)} \}$ are fixed points of
$\mathcal{M}$ or equivalently are critical points of $L$.
\end{theorem}

{\footnotesize
\begin{proof}
Let $\{ \lambda^{(k_n)} \} $ be a subsequence of
$\{ \lambda^{(k)} \}$ converging to $\bar\lambda$.
Then for all $\gamma \in
\Rn$:
\begin{eqnarray*}
A(\gamma , \lambda^{(k_n)} ) 
& \leq &
A(\hat\gamma^{(k_n)} , \lambda^{(k_n)} ) \quad \textrm{by definition
  of $\mathcal{M}$} \\
& \leq &
L(\hat\gamma^{(k_n)} + \lambda^{(k_n)}) - L(\lambda^{(k_n)}) \quad
\textrm{by Lemma \ref{A<L-L}} \\
& = &  
L( \lambda^{(k_n + 1)}) - L(\lambda^{(k_n)}) \quad \textrm{by
  definition of IM} \\
& \leq &
L( \lambda^{(k_{n+1})}) - L(\lambda^{(k_n)}) \quad \textrm{by
  monotonicity of $L(\lambda^{(k)})$,}
\end{eqnarray*}
and in the limit as $n \rightarrow \infty$, for continuous $A$ and $L$:
$A(\gamma , \bar\lambda) \leq L(\bar\lambda) -  L(\bar\lambda) = 0$.
Thus $\gamma = 0$ is a maximum of $A(\gamma , \bar\lambda)$,
using Lemma \ref{A0=0},
and $\bar\lambda$ is a fixed point of $\mathcal{M}$.
Furthermore, $\left. \frac{d}{dt} \right|_{t=0} A(t\gamma ,
\bar\lambda) = \left. \frac{d}{dt} \right|_{t=0} L(t\gamma +
\bar\lambda) = 0$,
using Lemma \ref{dA=dL},
and $\bar\lambda$ is a critical point of
$L$. 
\end{proof}
}
From this and Theorem \ref{IncreasingIMLikelihood} it follows
immediately that each sequence of likelihood values, for which an
upper bound exists, converges monotonically to a critical point of $L$.

\begin{corollary}
Let $\{ L (\lambda^{(k)} \}$ be a sequence of likelihood values
bounded from above. Then $\{ L (\lambda^{(k)} \}$ converges
monotonically to a value
$L^\ast = L(\lambda^\ast)$ for some critical point
$\lambda^\ast$ of $L$. 
\end{corollary}

\section*{Acknowledgements}

This work was supported by the Graduiertenkolleg ILS at the Seminar
f\"ur Sprachwissenschaft, T\"ubingen. The author would like to thank
Steve Abney, Mark Johnson, Graham Katz and Detlef Prescher for their
valuable comments on this paper.

\end{appendix}

}

\bibliographystyle{plain}

\end{document}